\documentclass[11pt,a4paper]{article}

\usepackage{authblk}
\usepackage[hyperref]{emnlp-ijcnlp-2019}
\usepackage{times}
\usepackage{latexsym}

\usepackage{rotating}
\usepackage[utf8]{inputenc}
\usepackage[ngerman]{babel}
\usepackage{graphicx}
\usepackage{array}
\usepackage{balance}
\usepackage{epsfig}
\usepackage{epstopdf}
\usepackage{subfigure}
\usepackage{multirow}
\usepackage{color}

\newcommand{\para}[1]{\medskip \noindent {\bf #1}}

\newenvironment{myquote}%
  {\list{}{\leftmargin=0.1in\rightmargin=0.1in}\item[]}%
  {\endlist}

\usepackage{tikz}

\usepackage{silence}
\usepackage{url}

\usepackage{arabtex}
\renewenvironment{abstract}%
         {\centerline{\large\bf Abstract}%
          \begin{list}{}%
             {\setlength{\rightmargin}{0.6cm}%
              \setlength{\leftmargin}{0.6cm}}%
           \item[]\ignorespaces}%
         {\unskip\end{list}}
         
\usepackage{tipa}
\usepackage{utf8}
\setcode{utf8}

\aclfinalcopy % Uncomment this line for the final submission

\title{Neural Arabic Text Diacritization: State of the Art Results and a Novel Approach for Machine Translation}

\author{Ali Fadel}
\author{Ibraheem Tuffaha}
\author{Bara' Al-Jawarneh}
\author{Mahmoud Al-Ayyoub}

\affil{Jordan University of Science and Technology, Irbid, Jordan \authorcr
  \{\tt aliosm1997, bro.t.1996, baraaaljawarneh, malayyoub\}@gmail.com}

\date{\today}

\begin{document}
\maketitle
\begin{abstract}
In this work, we present several deep learning models for the automatic diacritization of Arabic text. Our models are built using two main approaches, viz. Feed-Forward Neural Network (FFNN) and Recurrent Neural Network (RNN), with several enhancements such as 100-hot encoding, embeddings, Conditional Random Field (CRF) and Block-Normalized Gradient (BNG). The models are tested on the only freely available benchmark dataset and the results show that our models are either better or on par with other models, which require language-dependent post-processing steps, unlike ours. Moreover, we show that diacritics in Arabic can be used to enhance the models of NLP tasks such as Machine Translation (MT) by proposing the \emph{Translation over Diacritization} (ToD) approach.
\end{abstract}

\section{Introduction}

In Arabic and many other languages, diacritics are added to the characters of a word (as short vowels) in order to convey certain information about the meaning of the word as a whole and its place within the sentence. Arabic Text Diacritization (ATD) is an important problem with various applications such as text to speech (TTS). At the same time, this problem is a very challenging one even to native speakers of Arabic due to the many subtle issues in determining the correct diacritic for each character from the list shown in Figure~\ref{tab4_3} and the lack of practice for many native speakers. Thus, the need to build automatic Arabic text diacritizers is high \cite{zitouni2009arabic}.

The meaning of a sentence is greatly influenced by the diacritization which is determined by the context of the sentence as shown in the following example:

\begin{myquote}
\centering
\<
كلم أحمد ...
> \\
Buckwalter Transliteration: klm {\textgreater}Hmd ... \\
Incomplete sentence without diacritization. \\
\<
كَلَّمَ أَحْمَدٌ صَدِيقَهُ
>\\
Buckwalter Transliteration: k{\color{red}a}l{\color{red}$\sim$a}m{\color{red}a} {\textgreater}{\color{red}a}H{\color{red}o}m{\color{red}a}d{\color{red}N} S{\color{red}a}d{\color{red}i}yq{\color{red}a}h{\color{red}u} \\
Translation: Ahmad talked to his friend. \\
\<
كَلَمَ أَحْمَدٌ عَدُوَّهُ
> \\
Buckwalter Transliteration: k{\color{red}a}l{\color{red}a}m{\color{red}a} {\textgreater}{\color{red}a}H{\color{red}o}m{\color{red}a}d{\color{red}N} E{\color{red}a}d{\color{red}u}w{\color{red}$\sim$a}h{\color{red}u} \\
Translation: Ahmad wounded his enemy.
\end{myquote}

The letters
\<
كلم 
>
``klm'' manifests into two different words when given two different diacritizations. As shown in this example,
\<
كَلَّمَ 
>
``kal$\sim$ama'' in the first sentence is the verb `talked' in English, while
\<
كَلَمَ
>
``kalama'' in the second sentence is the verb `wounded' in English.

To formulate the problem in a formal manner: Given a sequence of characters representing an Arabic sentence $S$, find the correct diacritic class (from Figure~\ref{tab4_3}) for each Arabic character $S_i$ in $S$.

Despite the problem's importance, it received limited attention. One of the reasons for this is the scarcity of freely available resources for this problem. To address this issue, the Tashkeela Corpus\footnote{\url{https://sourceforge.net/projects/tashkeela}} \cite{zerrouki2017tashkeela} has been released to the community. Unfortunately, there are many problems with the use of this corpus for benchmarking purposes. A very recent study \cite{dataset} discussed in details these issues and provided a cleaned version of the dataset with pre-defined split into training, testing and validation sets. In this work, we use this dataset and provide yet another extension of it with a larger training set and a new testing set to circumvent the issue that some of the existing systems have already been trained on the entire Tashkeela Corpus.

According to \cite{dataset}, existing approaches to ATD are split into two groups: traditional rule-based approaches and machine learning based approaches. The former was the main approach by many researchers such as \cite{zitouni2009arabic,pasha2014madamira,darwish2017arabic} while the latter has started to receive attention only recently \cite{belinkov2015arabic,abandah2015automatic,shakkala,mubarak2019highly}. Based on the extensive experiments of \cite{dataset}, deep learning approaches (aka neural approaches) are superior to non-neural approaches especially when large training data is available. In this work, we present several neural ATD models and compare their performance with the state of the art (SOTA) approaches to show that our models are either on par with the SOTA approaches or even better.
Finally, we present a novel way to utilize diactritization in order to enhance the accuracy of Machine Translation (MT) models in what we call \emph{Translation over Diacritization} (ToD) approach.

The rest of the paper is organized as follows. The following section discusses the dataset proposed by \cite{dataset}.
Sections~\ref{sec:ffnn} and \ref{sec:rnn} discuss our two main approaches: Feed-Forward Neural Network (FFNN) and Recurrent Neural Network (RNN), respectively. Section~\ref{sec:comparison} brielfy discusses the related work and presents a comparison with the SOTA approaches while Section~\ref{sec:tod} describes our novel approach to integrate diacritization into translation tasks. The paper is concluding in Section~\ref{sec:conc} with final remarks and future directions of this work.

\section{Dataset}
\label{sec:dataset}

The dataset of \cite{dataset} (which is an adaptation of the Tashkeela Corpus) consists of about 2.3M words spread over 55K lines. Basic statistics about this dataset size, content and diacritics usage are given in Table~\ref{tab:dataset}.
Among the resources provided with this dataset are new definitions of the 
Diacritic Error Rate (DER), which is ``the percentage of misclassified Arabic characters regardless of whether the character has 0, 1 or 2 diacritics'', and the Word Error Rate (WER), which is ``the percentage of Arabic words which have at least one misclassified Arabic character''.\footnote{DER/WER are computed with \href{https://github.com/AliOsm/arabic-text-diacritization/blob/master/helpers/diacritization_stat.py}{diacritization\_stat.py}}
The redefinition of these measures is to exclude counting irrelevant characters such as numbers and punctuations, which were included in \cite{zitouni2009arabic}'s original definitions of DER and WER.
It is worth mentioning that DER/WER are computed in four different ways in the literature depending on whether the last character of each word (referred to as case ending) is counted or not and whether the characters with no diacritization are counter or not.

\begin{table}
\centering
\caption{Statistics about the size, content and diacritics usage
of \cite{dataset}'s Dataset}
\label{tab:dataset}
\begin{tabular}{c|c|c|c|}
\cline{2-4}
\multicolumn{1}{l|}{}                   & Train & Valid & Test \\ \hline
\multicolumn{1}{|c|}{Words Count}       & 2,103K   & 102K  & 107K    \\ \hline
\multicolumn{1}{|c|}{Lines Count}       & 50K      & 2.5K  & 2.5K    \\ \hline
\multicolumn{1}{|c|}{Avg Chars/Word}    & 3.97     & 3.97  & 3.97    \\ \hline
\multicolumn{1}{|c|}{Avg Words/Line}    & 42.06    & 40.97 & 42.89   \\ \hline
\multicolumn{1}{|c|}{0 Diacritics (\%)} & 17.78    & 17.75 & 17.80   \\ \hline
\multicolumn{1}{|c|}{1 Diacritic (\%)}  & 77.17    & 77.19 & 77.22   \\ \hline
\multicolumn{1}{|c|}{2 Diacritics (\%)} & 5.03     & 5.05  & 4.97    \\ \hline
\multicolumn{1}{|c|}{Error Diacritics (\%)} & 0 & 0 & 0              \\ \hline
\end{tabular}
\end{table}

\section{The Feed-Forward Neural Network (FFNN) Approach}
\label{sec:ffnn}

This is our first approach and we present three models based on it.
In this approach, we consider diacritizing each character as an independent problem. To do so, the model takes a 100-dimensional vector as an input representing features for a single character in the sentence. The first 50 elements in the vector represent the 50 non-diacritic characters before the current character and the last 50 elements represent the 50 non-diacritic characters after it including the current character.

\begin{figure}[h]
    \centering
    \includegraphics[width=0.48\textwidth]{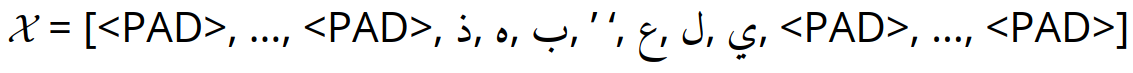}
    \caption{Vector representation of a FFNN example.}
    \label{vector}
\end{figure}

For example, the sentence `\<
ذَهَبَ عَلِي
>', the vector related to the character
`\<
ب
>'
is as shown in Figure~\ref{vector}.
As the figure shows,
there are two characters before the character
`\<
ب
>'
and four after it (including the whitespace). The special token `$<$PAD$>$' is used as a filler when there are no characters to feed. Note that the dataset contains 73 unique characters (without the diacritics) which are mapped to unique integer values from 0 to 74 after sorting them based on their unicode representations including the special padding and unknown (`$<$UNK$>$') tokens.

\begin{figure}
    \centering
    \includegraphics[width=0.48\textwidth]{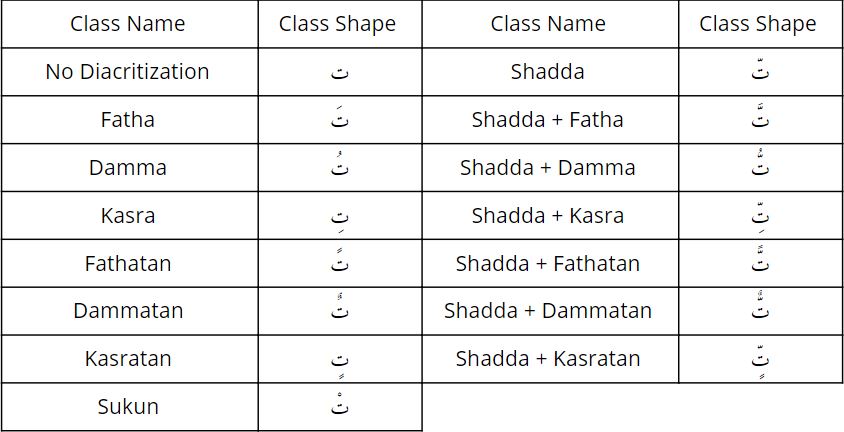}
    \caption{The 15 classes under consideration.}
    \label{tab4_3}
\end{figure}

Each example belongs to one of the 15 classes under consideration, which are shown in Figure~\ref{tab4_3}.
The model outputs probabilities for each class. Using a Softmax output unit, the class with maximum probability is considered as the correct output.
The number of training, validation and testing examples from converting the dataset into examples as described earlier are 9,017K, 488K and 488K respectively.

\para{Basic Model.}
The basic model consists of 17 hidden layers of different sizes. The activation function used in all layers is Rectified Linear Unit (ReLU) and the number of trainable parameters is about 1.5M. For more details see Appendix~\ref{app:a}.
The model is trained for 300 epochs on an Nvidia GeForce GTX 970M GPU for about 16 hours using AdaGrad optimization algorithm \cite{duchi2011adaptive} with 0.01 learning rate, 512 batch size, and categorical cross-entropy loss function.

\para{100-Hot Model.}
In this model, each integer from the 100-integer inputs is converted into its 1-hot representation as a 75-dimensional vector. Then, the 100 vectors are concatenated forming a 7,500-dimensional vector.
Based on empirical exploration, the model is structured to have five hidden layers with dropout. It has close to 2M trainable parameters. For more details see Appendix~\ref{app:a}.
The model is trained for 50 epochs on an Nvidia GeForce GTX 970M GPU for about 3 hours using Adam optimization algorithm \cite{kingma2014adam} with 0.001 learning rate, 0.9 beta1, 0.999 beta2, 512 batch size, and categorical cross-entropy loss function.

\para{Embeddings Model.}
In this model, the 100-hot layer is replaced with an embeddings layer to learn feature vectors for each character through the training process.
Empirically determined, the model has five hidden layers with only 728K trainable parameters.
For more details see Appendix~\ref{app:a}.
The model is trained with the same configurations as the 100-hot model and the training time is about 2.5 hours only.

\para{Results and Analysis.}
Although the idea of diacritizing each character independently is counter-intuitive, the results of the FFNN models on the test set (shown in Table~\ref{tab:ffnnderwer}) are very promising with the embeddings model having an obvious advantage over the basic and 100-hot models and performing much better than the best rule-based diacritization system Mishkal\footnote{\url{https://tahadz.com/mishkal}} among the systems reviewed by \cite{dataset} (Mishakl DER: 13.78\% vs FFNN Embeddings model DER: 4.06\%). However, these models are still imperfect.
More detailed error analysis of these models is available in Appendix~\ref{app:a}.

\begin{table*}
\centering
\caption{DER/WER comparison of the different FFNN models on the test set}
\label{tab:ffnnderwer}
\begin{tabular}{|c|c|c|c|c|}
\hline
\multirow{2}{*}{DER/WER} & w/ case ending & w/o case ending & w/ case ending & w/o case ending \\ \cline{2-5} 
 & \multicolumn{2}{c|}{Including `no diacritic'} & \multicolumn{2}{c|}{Excluding `no diacritic'} \\ \hline
Basic model & \small9.33\% / \small25.93\% & \small6.58\% / \small13.89\% & \small10.85\% / \small25.39\% & \small7.51\% / \small13.53\% \\ \hline
100-Hot model & \small6.57\% / \small20.21\% & \small4.83\% / \small11.14\% & \small7.75\% / \small19.83\% & \small5.62\% / \small10.93\% \\ \hline
Embeddings model & \textbf{\small5.52\% / \small17.12\%} & \textbf{\small4.06\% / \small9.38\%} & \textbf{\small6.44\% / \small16.63\%} & \textbf{\small4.67\% / \small9.10\%} \\ \hline
\end{tabular}
\end{table*}

\section{The Recurrent Neural Network (RNN) Approach}
\label{sec:rnn}

Since RNN models usually need huge data to train on and learn high-level linguistic abstractions, we prepare an external training dataset following the guidelines of \cite{dataset}. The extra training dataset is extracted from the Classical Arabic (CA) part of the Tashkeela Corpus and the Holy Quran (HQ). We exclude the lines that already exist in the previously mentioned dataset. Note that, with the extra training dataset the number of unique characters goes up to 87 (without the diacritics). Table~\ref{tab:extra_dataset} shows the statistics for the extra training dataset.

\begin{table}
\centering
\caption{Extra training dataset statistics}
\label{tab:extra_dataset}
\begin{tabular}{c|c|}
\cline{2-2}
                                            & Extra Train \\ \hline
\multicolumn{1}{|c|}{Words Count}           & 22.4M       \\ \hline
\multicolumn{1}{|c|}{Lines Count}           & 533K        \\ \hline
\multicolumn{1}{|c|}{Avg Chars/Word}        & 3.97        \\ \hline
\multicolumn{1}{|c|}{Avg Words/Line}        & 42.1        \\ \hline
\multicolumn{1}{|c|}{0 Diacritics (\%)}     & 17.79       \\ \hline
\multicolumn{1}{|c|}{1 Diacritic (\%)}      & 77.16       \\ \hline
\multicolumn{1}{|c|}{2 Diacritics (\%)}     & 5.03        \\ \hline
\multicolumn{1}{|c|}{Error Diacritics (\%)} & 0           \\ \hline
\end{tabular}
\end{table}

The lines in the dataset are split using the following 14 punctuations (`.', `,', 
`\<،>',
`:', `;', 
`\<؛>'
, `(', `)', `[', `]', `\{', `\}', `«' and `»'). After that, the lines with length more than 500 characters (without counting diacritics) are split into lines of length no more than 500. This step is necessary for the training phase to limit memory usage within a single batch. Note that the splitting procedure is omitted within the prediction phase, e.g., when calculating DER/WER on the validation and test sets. Moreover, four special tokens (`$<$SOS$>$', `$<$EOS$>$', `$<$UNK$>$' and `$<$PAD$>$') are used to prepare the input data before feeding it to the model. `$<$SOS$>$' and `$<$EOS$>$' are added to the start and the end of the sequences, respectively. `$<$UNK$>$' is used to represent unknown characters not seen in the training dataset. Finally, `$<$PAD$>$' is appended to pad the sequences within the same batch. Four equivalent special tokens are used as an output in the target sequences.

\para{Basic Model.}
Several model architectures are trained without the extra training dataset. After some exploration, the best model architecture is chosen to experiment with different techniques as described in details throughout this section.

\begin{figure}[t]
    \centering
    \includegraphics[width=0.48\textwidth]{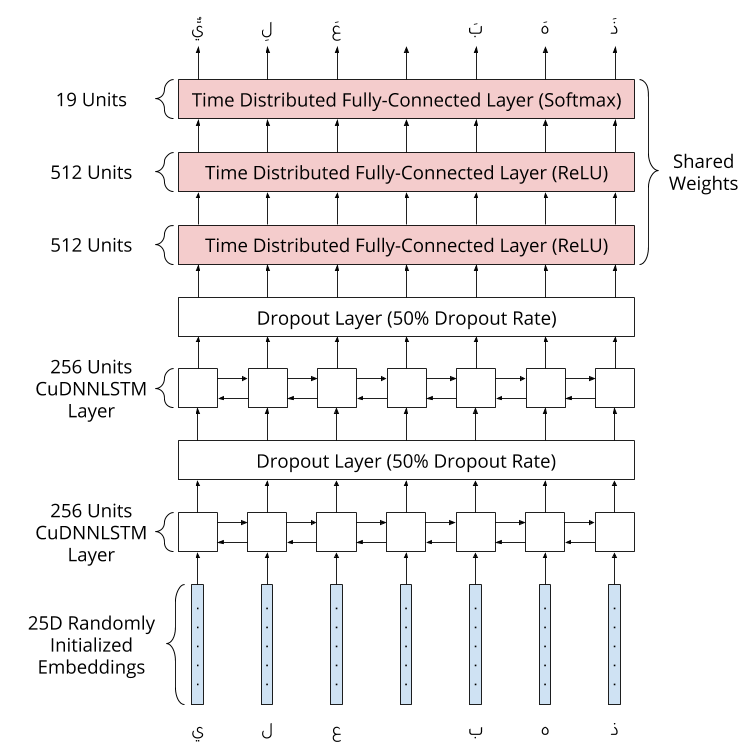}
    \caption{RNN basic model structure.}
    \label{fig:rnn_basic}
\end{figure}

The exploration is done to tune different hyperparameters and find the structure that gives the best DER, which, in most cases, leads to better WER. Because the neural network size have a great impact on performance, we primarily experiment with the number of Bidirectional CuDNN Long Short-Term Memory (BiCuDNNLSTM) \cite{appleyard2016optimizing} layers and their hidden units. By using either one, two or three layers, the error significantly decreases going from one layer to two layers. However, it shows slight improvement (if any) when going from two layers to three layers while increasing the training time. So, we decide to use two BiCuDNNLSTMs in further experiments as well as 256 hidden units per layer as using less units will increase the error rate while using more units does not significantly improve it.
Then, we experiment with the size and depth of the fully connected feed-forward network. The results show that the depth is not as important as the size of each layer. The best results are produced with the model using two layers with 512 hidden units each.
All experiments are done using Adam optimization algorithm, because different optimizers like Stochastic Gradient Descent, Adagrad and Adadelta do not converge to the optimal minimal fast enough and RMSprop, Nadam and Adamax give the same or slightly worse results.
The number of character features to learn in the embedding layer that gives the best results is 25, where more features leads to little improvement and more overfitting, and less features makes the training harder for the network. This is probably due to the input vocabulary being limited to 87 different characters.
We also experiment with training the models for more than 50 epochs, but the return is very little or it makes the learning unstable and eventually causes exploding gradients leaving the network with useless predictions, unable to learn anymore.
The best model is structured as shown in
Figure~\ref{fig:rnn_basic}.

The training is done twice: with and without the extra training dataset, in order to explore the impact of the dataset size on the training phase for the diacritization problem. This has led to reduced overfitting. A weights averaging technique over the last few epochs is applied to partially overcome the overfitting issue and obtain a better generalization.

Models in all following experiments are trained on Google Colab\footnote{\url{http://colab.research.google.com}} \cite{carneiro2018performance} environment for 50 epochs using an Nvidia Tesla T4 GPU, Adam  optimization algorithm with 0.001 learning rate, 0.9 beta1, 0.999 beta2, $10^{-7}$ epsilon, 256 batch size, and categorical cross-entropy loss function.

\para{Conditional Random Field (CRF) Model.}
A CRF classifier is used in this model instead of the Softmax layer to predict the network output. CRF is usually more powerful than Softmax in terms of sequence dependencies in the output layer which exist in the diacritization problem.
It is worth mentioning that CRF is considered to be ``a best practice'' in sequence labeling problems. However, in this particular problem, the results show that CRF performs worse than Softmax in most cases except for WER results when training without the extra dataset which indicates that, even with worse DER results, CRF is able to make more consistent predictions within the same word.

\para{Block-Normalized Gradient (BNG) Model.}
In this model, \cite{yu2017block}'s BNG method is applied to normalize gradients within each batch. This can help accelerate the training process. According to \cite{yu2017block}, this method performs better in RNN when using optimizers with adaptive step sizes, such as Adam. It can also lead to solutions with better generalization. This coincides with our results.
%compared to the results of the basic and CRF models.

\begin{table*}[h]
\centering
\caption{DER/WER comparison of the different RNN models on the test set}
\label{tab:rnn_derwer}
\begin{tabular}{|c|c|c|c|c|}
\hline
\multirow{2}{*}{DER/WER} & w/ case ending         & w/o case ending        & w/ case ending         & w/o case ending        \\ \cline{2-5} 
                         & \multicolumn{2}{c|}{Including `no diacritic'}   & \multicolumn{2}{c|}{Excluding `no diacritic'}   \\ \hline
\multicolumn{5}{|c|}{Without Extra Train Dataset}                                                                            \\ \hline
Basic model              & \small2.68\% / \small7.91\%          & \small2.19\% / \small4.79\%          & \small3.09\% / \small7.61\%          & \small2.51\% / \small4.66\%          \\ \hline
CRF model                & \small2.67\% / \small7.73\%          & \small2.19\% / \small4.69\%          & \small3.08\% / \small7.46\%          & \small2.52\% / \small4.60\%           \\ \hline
BNG model                & \textbf{\small2.60\% / \small7.69\%}  & \textbf{\small2.11\% / \small4.57\%} & \textbf{\small3.00\% / \small7.39\%} & \textbf{\small2.42\% / \small4.44\%} \\ \hline
\multicolumn{5}{|c|}{With Extra Train Dataset}                                                                               \\ \hline
Basic model              & \small1.72\% / \small5.16\%          & \small1.37\% / \small2.98\%          & \small1.99\% / \small4.96\%          & \small1.59\% / \small2.92\%          \\ \hline
CRF model                & \small1.84\% / \small5.42\%          & \small1.47\% / \small3.17\%          & \small2.13\% / \small5.22\%          & \small1.69\% / \small3.09\%          \\ \hline
BNG model                & \textbf{\small1.69\% / \small5.09\%} & \textbf{\small1.34\% / 2.91\%} & \textbf{\small1.95\% / \small4.89\%} & \textbf{\small1.54\% / \small2.83\%} \\ \hline
\end{tabular}
\end{table*}

\begin{table*}[h]
\centering
\caption{DER/WER comparison showing the effect of the weights averaging technique on BNG model}
\label{tab:rnn_bng_derwer}
\begin{tabular}{|c|c|c|c|c|c|}
\hline
\multirow{2}{*}{\begin{tabular}[c]{@{}c@{}}DER/WER\end{tabular}} & \multirow{2}{*}{\begin{tabular}[c]{@{}c@{}}Averaged\\ Epochs\end{tabular}} & w/ case ending & w/o case ending & w/ case ending & w/o case ending \\ \cline{3-6} 
 &  & \multicolumn{2}{c|}{Including `no diacritic'} & \multicolumn{2}{c|}{Excluding `no diacritic'} \\ \hline
\multirow{4}{*}{\begin{tabular}[c]{@{}c@{}}Without\\ extra\\ train\\ dataset\end{tabular}} & 1 & \small2.73\% / \small8.08\% & \small2.21\% / \small4.80\% & \small3.16\% / \small7.79\% & \small2.54\% / \small4.68\% \\ \cline{2-6} 
 & 5 & \small2.64\% / \small7.80\% & \small2.14\% / \small4.64\% & \small3.04\% / \small7.49\% & \small2.46\% / \small4.52\% \\ \cline{2-6} 
 & 10 & \textbf{\small2.60\% / \small7.69\%} & \textbf{\small2.11\%} / \small4.57\% & \textbf{\small3.00\%/ \small7.39\%} & \textbf{\small2.42\% / \small4.44\%} \\ \cline{2-6} 
 & 20 & \small2.61\% / \small7.73\% & \textbf{\small2.11\% / \small4.56\%} & \small3.01\% / \small7.42\% & \textbf{\small2.42\%} / \small7.42\% \\ \hline
\multirow{4}{*}{\begin{tabular}[c]{@{}c@{}}With\\ extra\\ train\\ dataset\end{tabular}} & 1 & \small1.97\% / \small5.85\% & \small1.61\% / \small3.55\% & \small2.20\% / \small5.61\% & \small1.82\% / \small3.45\% \\ \cline{2-6} 
 & 5 & \small1.73\% / \small5.20\% & \small1.38\% / \small3.02\% & \small1.98\% / \small4.98\% & \small1.58\% / \small2.92\% \\ \cline{2-6} 
 & 10 & \small1.70\% / \small5.13\% & \small1.35\% / \small2.94\% & \small1.96\% / \small4.92\% & \small1.55\% / \small2.85\% \\ \cline{2-6} 
 & 20 & \textbf{\small1.69\% / \small5.09\%} & \textbf{\small1.34\% / \small2.91\%} & \textbf{\small1.95\% / \small4.89\%} & \textbf{\small1.54\% / \small2.83\%} \\ \hline
\end{tabular}
\end{table*}

\para{Discussion and Analysis.}
The results of the RNN models on the test set (shown in Table~\ref{tab:rnn_derwer}) are much better than the FFNN models by about 67\%. To show the effect of the weights averaging technique, Table~\ref{tab:rnn_bng_derwer} reports the DER/WER statistics related to the BNG model after averaging its weights over the last 1, 5, 10, and 20 epochs.
Studying the confusion matrices for all the models suggests that the Shadda class and the composite classes (i.e., Shadda + another diacritic) are harder to learn for the network compared to other classes. However, with the extra training dataset, the network is able to find significantly better results compared to the results without the extra training dataset, especially for the Shadda class.

The comparison method for calculating DER/WER without case ending skips comparing the diacritization on the end of each word. This skip improves the best DER to 1.34\% (vs 1.69\%) and best WER to 2.91\% (vs 5.09\%) which is a 26\% improvement in DER and 43\% improvement in WER. This is because the diacritic of the last character of the word usually depends on the part of speech tag making it harder to diacritize. However, we note that the actual last character of the word may come before the end of the word if the word has some suffix added to it.

\begin{table*}[h]
\centering
\caption{Comparing the BNG model with \cite{shakkala} in terms of DER/WER on the test set}
\label{tab:shakkala_derwer}
\makebox[\textwidth]{%
\begin{tabular}{|c|c|c|c|c|}
\hline
\multirow{2}{*}{DER/WER} & w/ case ending & w/o case ending & w/ case ending & w/o case ending \\ \cline{2-5} 
 & \multicolumn{2}{c|}{Including `no diacritic'} & \multicolumn{2}{c|}{Excluding `no diacritic'} \\ \hline
\multicolumn{5}{|c|}{\cite{dataset} Testing Dataset Results} \\ \hline
Our best model & \textbf{\small1.78\% / \small5.38\%} & \textbf{\small1.39\% / \small3.04\%} & \textbf{\small2.05\% / \small5.17\%} & \textbf{\small1.60\% / \small2.96\%} \\ \hline
Barqawi, 2017 & \small3.73\% / \small11.19\% & \small2.88\% / \small6.53\% & \small4.36\% / \small10.89\% & \small3.33\% / \small6.37\% \\ \hline
\multicolumn{5}{|c|}{Auxiliary Testing Dataset Results} \\ \hline
Our best model & \textbf{\small5.98\% / \small15.72\%} & \small5.21\% / \small11.07\% & \textbf{\small5.54\% / \small13.21\%} & \textbf{\small4.85\% / \small9.02\%} \\ \hline
Barqawi, 2017 & \small6.41\% / \small17.52\% & \textbf{\small5.12\% / \small10.91\%} & \small6.82\% / \small15.92\% & \small5.32\% / \small9.65\% \\ \hline
\end{tabular}}
\end{table*}

\begin{table*}[h]
\centering
\caption{Comparing the BNG model with \cite{belinkov2015arabic} in terms of DER/WER on the test set}
\label{tab:belinkov_derwer_original_testset}
\makebox[\textwidth]{%
\begin{tabular}{|c|c|c|c|c|}
\hline
\multirow{2}{*}{DER/WER} & w/ case ending & w/o case ending & w/ case ending & w/o case ending \\ \cline{2-5} 
 & \multicolumn{2}{c|}{Including `no diacritic'} & \multicolumn{2}{c|}{Excluding `no diacritic'} \\ \hline
\multicolumn{5}{|c|}{Classical Arabic Testing Dataset Results} \\ \hline
Our best model & \textbf{\small1.99\% / \small6.10\%} & \textbf{\small1.48\% / \small3.25\%} & \textbf{\small2.30\% / \small5.88\%} & \textbf{\small1.70\% / \small3.17\%} \\ \hline
Belinkov, 2015 & \small31.26\% / \small75.29\% & \small29.66\% / \small59.46\% & \small35.78\% / \small74.37\% & \small33.67\% / \small57.66\% \\ \hline
\multicolumn{5}{|c|}{Modern Standard Arabic Testing Dataset Results} \\ \hline
Our best model & \textbf{\small8.05\% / \small23.56\%} & \textbf{\small6.85\% / \small16.12\%} & \textbf{\small8.29\% / \small21.10\%} & \textbf{\small7.16\% / \small14.41\%} \\ \hline
Belinkov, 2015 & \small31.77\% / \small75.02\% & \small29.21\% / \small59.40\% & \small37.13\% / \small73.93\% & \small33.82\% / \small58.03\% \\ \hline
\end{tabular}}
\end{table*}

\begin{figure}[h]
    \centering
    \includegraphics[width=0.48\textwidth]{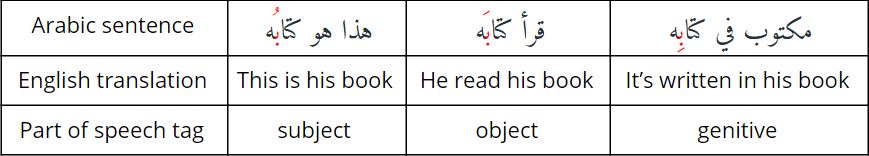}
    \caption{Case ending different diacritization with different part of speech tag.}
    \label{fig:pos}
\end{figure}

Consider the example shown in Figure~\ref{fig:pos}. The word
`\<كتابه’>'
means `his book' where the last character
`\<ـه’>'
is the suffix representing the pronoun `his', and the letter before it may take three different diacritics depending on its part of speech tagging.
More detailed error analysis of these models available in Appendix~\ref{app:b}.

Furthermore, an Encoder-Decoder structure (seq2seq) was built using BiCuDNNLSTMs to encode a sequence of characters and generate a sequence of diacritics, but the model was not able to successfully learn the alignment between inputted characters and outputted diacritics. Other attempts tried encoding the sentences as sequences of words and generate a sequences of diacritics also terribly failed to learn.

The BNG model performs the best compared to other models described above. So, it is used for comparison with other systems in the following section.

\section{Comparison with Existing Systems}
\label{sec:comparison}

As mentioned earlier, the efforts on building automatic ATD is limited. A recent study \cite{dataset} surveyed existing approaches and tools for ATD. After discussing the limitations in closed-source tools, they divided existing approaches to ATD into two groups:
traditional rule-based approaches \cite{zitouni2009arabic,pasha2014madamira,shahrour2015improving,alnefaie2017automatic,bebah2014hybrid,azmi2015survey,chennoufi2017morphological,darwish2017arabic,fashwan2017shakkil,alqahtani2019homograph}
and machine learning based approaches \cite{belinkov2015arabic,abandah2015automatic,abandah2017investigating,shakkala,moumen2018evaluation,mubarak2019highly}. The extensive experiments of \cite{dataset} showed that neural ATD models are superior to their competitors especially when large training data is available. Thus, we limit our attention in this work to such models.

According to \cite{dataset}, the Shakkala system \cite{shakkala} performs the best compared to other existing systems using the test set and the evaluation metrics proposed in \cite{dataset}. Considering our best model's results mentioned previously, it is clear that our model outperforms Shakkala on the testing set after splitting the lines to be at most 315 characters long (Shakkala system limit), which causes a slight drop in our best model's results. However, since Shakkala was also trained on Tashkeela Corpus, we develop an auxiliary test set extracted from three books from Al-Shamela Library\footnote{\url{http://shamela.ws}}
`\<تاج العروس من جواهر القاموس>', `\<الفتاوى الكبرى لابن تيمية>' and `\<فتح الباري شرح صحيح البخاري>'
using the same extraction and cleaning method proposed by \cite{dataset} while keeping only lines with more than 80\% ``diacritics to Arabic characters'' rate. The extracted lines are each split into lines of lengths no more than 315 characters (without counting diacritics) which is the input limit of the Shakkala system. This produces a test set consisting of 443K words. Table~\ref{tab:shakkala_derwer} shows the results comparison with Shakkala.

A comparison with the pre-trained model of \cite{belinkov2015arabic} is also done using the test set and the evaluation metrics of \cite{dataset} while splitting the lines into lines of lengths no more than 125 characters (without counting diacritics) since any input with length more than that causes an error in their system. The results show that \cite{belinkov2015arabic}'s model performs poorly.
However, we note that \cite{belinkov2015arabic}'s system  was trained and tested on the Arabic TreeBank (ATB) dataset which consists of text in Modern Standard Arabic (MSA). So, to make a fair comparison with \cite{belinkov2015arabic}'s system, an auxiliary dataset is built from the MSA part of the Tashkeela Corpus using the same extraction and cleaning method proposed by \cite{dataset} keeping only lines with more than 80\% ``diacritics to Arabic characters'' rate. This test set consists of 111K words. The results are reported in Table~\ref{tab:belinkov_derwer_original_testset}.
In addition to the poor results of \cite{belinkov2015arabic}'s system, its output has a large number of special characters inserted randomly. These characters are removed manually to make the evaluation of the system possible.

Finally, we compare our model with \cite{abandah2015automatic}'s model which, to our best knowledge, is the most recent deep-learning work announcing the best results so far. To do so, we employ a similar comparison method to \cite{chennoufi2017morphological}'s by using the 10 books from the Tashkeela Corpus and the HQ
that were excluded from \cite{abandah2015automatic}'s test set. The sentences used for testing our best model are all sentences that are not included in the training dataset of \cite{dataset} or extra training dataset on which our model is trained. To make the comparison fair, we use the same evaluation metric as \cite{abandah2015automatic}, which is \cite{zitouni2009arabic}'s. Moreover, the characters with no diacritics in the original text are skipped similarly to \cite{abandah2015automatic}. The results are shown in Table~\ref{tab:abandah}.
It is worth mentioning that the results of \cite{abandah2015automatic} include post-processing techniques, which improved DER by 23.8\% as reported in \cite{abandah2015automatic}. It can be easily shown that, without this step, our model's results are actually superior.

\begin{table*}[h]
\centering
\caption{Comparing the BNG model with \cite{abandah2015automatic} in terms of DER/WER on the test set}
\label{tab:abandah}
\begin{tabular}{c|c|c|c|c|}
\cline{2-5}
\multirow{2}{*}{} & \multicolumn{2}{c|}{DER} & \multicolumn{2}{c|}{WER} \\ \cline{2-5} 
 & \begin{tabular}[c]{@{}c@{}}w/ case ending\end{tabular} & \begin{tabular}[c]{@{}c@{}}w/o case ending\end{tabular} & \begin{tabular}[c]{@{}c@{}}w/ case ending\end{tabular} & \begin{tabular}[c]{@{}c@{}}w/o case ending\end{tabular} \\ \hline
\multicolumn{1}{|c|}{\begin{tabular}[c]{@{}c@{}}Our best model\end{tabular}} & \small2.18\% & \small1.76\% & \textbf{\small4.44\%} & \textbf{\small2.66\%} \\ \hline
\multicolumn{1}{|c|}{\begin{tabular}[c]{@{}c@{}}Abandah, 2015\end{tabular}} & \textbf{\small2.09\%} & \textbf{\small1.28\%} & \small5.82\% & \small3.54\% \\ \hline
\end{tabular}
\end{table*}

All codes related to the diacritization work are publicly available on GitHub,\footnote{\url{https://github.com/AliOsm/shakkelha}} and are also implemented into a web application\footnote{\url{https://shakkelha.herokuapp.com}} for testing purposes.

\section{Translation over Diacritization (ToD)}
\label{sec:tod}

Word's diacritics can carry various types of information about the word itself, like its part of speech tag, the semantic meaning and the pronunciation. Intuitively, providing such extra features in NLP tasks has the potential to improve the results of any system. In this section, we show how we benefit from the integration of diacritics into Arabic-English (Ar-En) Neural Machine Translation (NMT) creating what we call Translation over Diacritization (ToD).

\para{Dataset Extraction and Preparation.}
Due to the lack of free standardized benchmark datasets for Ar-En MT, we create a mid-size dataset using the following corpora: GlobalVoices v2017q3, MultiUN v1, News-Commentary v11, Tatoeba v2, TED2013 v1.1, Ubuntu v14.10, Wikipedia v1.0~\cite{tiedemann2012parallel} downloaded from the OPUS\footnote{\url{http://opus.nlpl.eu}} project. The dataset contains 1M Ar-En sentence pairs split into 990K pairs for training and 10K pairs for testing. The extracted 1M pairs follow these conventions: (i) The maximum length for each sentence in the pair is 50 tokens, (ii) Arabic sentences contain Arabic letters only, (iii) English sentences contain English letters only, and (iv) the sentences do not contain any URLs.

The Arabic sentences in the training and testing datasets are diacritized using the best BNG model. After that, Byte Pair Encoding (BPE)\footnote{\url{https://github.com/rsennrich/subword-nmt}} \cite{sennrich2015neural} is applied separately on both English and original (undiacritized) Arabic sequences to segment the words into subwords. This step overcomes the Out Of Vocabulary (OOV) problem and reduces the vocabulary size. Then, diacritics are added to Arabic subwords to create the diacritized version. Table~\ref{tab:vocab_size} shows the number of tokens before and after BPE step for English, Original Arabic and Diacritized Arabic as well as the Diacritics forms when removing the Arabic characters.

\begin{table}[t]
\centering
\caption{Vocab size for all sequences types before and after BPE step}
\label{tab:vocab_size}
\begin{tabular}{|c|c|c|}
\hline
\multirow{2}{*}{Language} & \multicolumn{2}{c|}{Vocab Size} \\ \cline{2-3} 
                          & Before BPE      & After BPE     \\ \hline
English                   & 113K            & 31K           \\ \hline
Original Arabic                    & 224K            & 32K           \\ \hline
Diacritized Arabic           & 402K            & 186K          \\ \hline
Diacritics Forms          & 41K             & 15K           \\ \hline
\end{tabular}
\end{table}

\begin{figure}[h]
    \centering
    \includegraphics[width=0.48\textwidth]{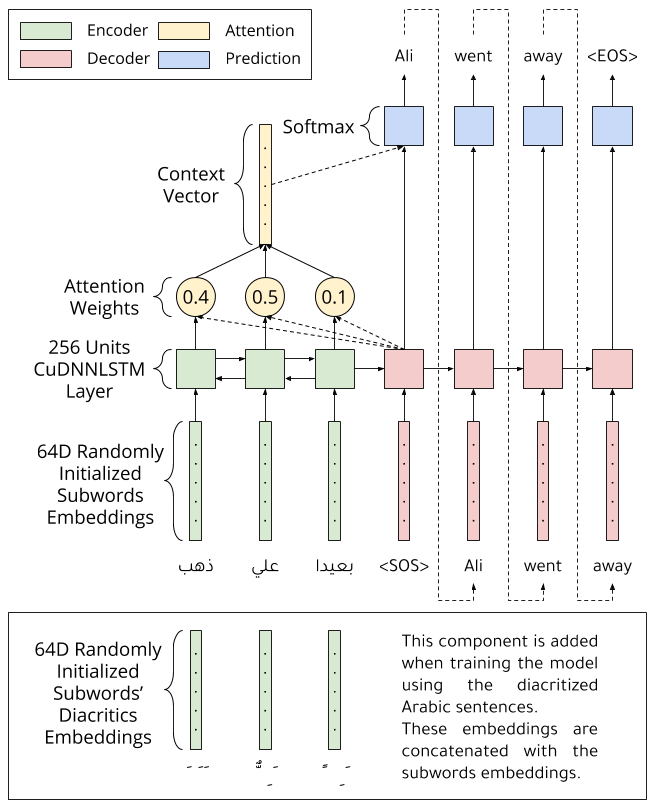}
    \caption{ToD model structure.}
    \label{fig:tod_structure}
\end{figure}

\para{Model Structure}
The model used in the experiments is a basic Encoder-Decoder sequence to sequence (seq2seq) model that consists of a BiCuDNNLSTM layer for encoding and a CuDNNLSTM layer for decoding with 512 units each (256 per direction for the encoder) while applying additive attention \cite{bahdanau2014neural} on the outputs of the encoder. As for the embeddings layer, a single randomly initialized embeddings layer with vector size 64 is used to represent the subwords when training without diacritics. Another layer with the same configuration is used to represent subwords' diacritics, which is concatenated with the subwords embeddings when training with diacritics. The model structure shown in Figure~\ref{fig:tod_structure}.

\para{Results and Discussion}
To explore the effect of the Arabic diacritization on the NMT task, we experiment with training both with and without diacritics. The models are trained for 50 epochs using an Nvidia Titan Xp GPU, Adam optimization algorithm with 0.001 learning rate, 0.9 beta1, 0.999 beta2, $10^{-7}$ epsilon and 256 batch size.

The structure for training the model with diacritics may vary. We experiment with two variations where the first one uses the diacritized version of the sequences, while the other one uses the original sequences and the diacritics sequences in parallel. When merging diacritics with their sequences, we get more variations of each word depending on its different forms of diacritization, therefore expanding the vocabulary size. On the other hand, when separating diacritics from their sequences, the vocab size stays the same, and diacritics are added separately as extra input.

\begin{figure}[h]
    \centering
    \includegraphics[width=0.48\textwidth]{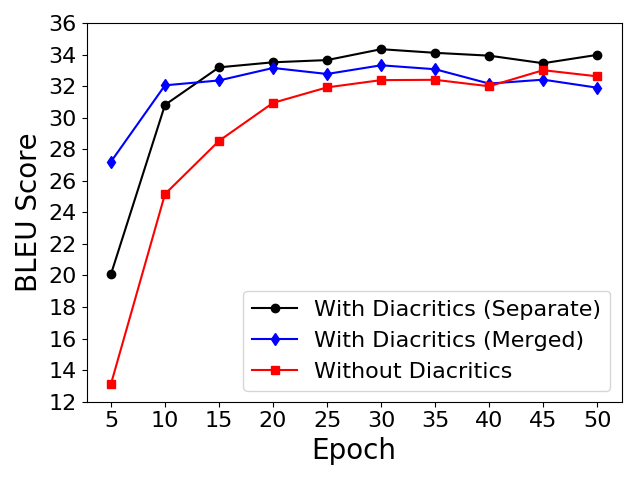}
    \caption{Testing dataset BLEU score while training.}
    \label{fig:bleu_scores}
\end{figure}

The results in Table~\ref{tab:tod_results} show that training the model with diacritization compared to without diacritization improves marginally by 0.31 BLEU score\footnote{BLEU scores are computed with \href{https://github.com/moses-smt/mosesdecoder/blob/master/scripts/generic/multi-bleu.perl}{multi-bleu.perl}}
when using the `with diacritics (merged)' data and improves even more when using the `with diacritics (separated)' data by 1.33 BLEU score. Moreover, the training time and model size increases by about 20.6\% and 41.4\%, respectively, for using the `with diacritics (merged)' data, while they only increase by about 3.4\% and 4.5\%, respectively, for using the `with diacritics (separated)' data. By observing Figure~\ref{fig:bleu_scores}, which reports the BLEU score on all three models every 5 epochs, it is clear that, although the `with diacritics (merged)' model converges better at the start of the training, it starts diverging after 15 epochs, which might be due to the huge vocab size and the training data size.

By analysing Figure~\ref{fig:bleu_scores}, we find that BLUE score converges faster when training with diacritics (merged) compared to the other two approaches. However, it starts diverging later on due to vocabulary sparsity. As for with diacritics (separated), the BLUE score has higher convergence compared to without diacritics while also maintaining stability compared to with diacritics (merged). This is because separating diacritics solves the vocabulary sparsity issue while also providing the information needed to disambiguate homonym words.

We note that, concurrently to our work, another work on utilizing diacritization for MT has recently appeared.
\cite{alqahtani2019homograph} used diacritics with text in three downstream tasks, namely Semantic Text Similarity (STS), NMT and Part of Speech (POS) tagging, to boost the performance of their systems. They applied different techniques to disambiguate homonym words through diacritization. They achieved 27.1 and 27.3 BLUE scores without and with diacritics, respectively, using their best disambiguation technique. This is a very small improvement of 0.74\% compared to our noticeable improvement of 4.03\%. Moreover, our approach is simpler and it does not require to drop any diacritical information.

All codes related to the ToD work are publicly available on GitHub\footnote{\url{https://github.com/AliOsm/translation-over-diacritization}}.

\begin{table}[t]
\centering
\caption{Translation over Diacritization (ToD) results on the test set}
\label{tab:tod_results}
\begin{tabular}{|c|c|c|c|}
\hline
Model & \begin{tabular}[c]{@{}c@{}}Training\\ Time\end{tabular} & \begin{tabular}[c]{@{}c@{}}Model\\ Size\end{tabular} & \begin{tabular}[c]{@{}c@{}}Best BLEU\\ Score\end{tabular} \\ \hline
Without & \textbf{29 Hours} & \textbf{285MB} & 33.01 \\ \hline
Merged & 35 Hours & 403MB & 33.32 \\ \hline
Separated & 30 Hours & 298MB & \textbf{34.34} \\ \hline
\end{tabular}
\end{table}

\section{Conclusion}
\label{sec:conc}
In this work, we explored the ATD problem. Our models, which follow two main approaches: FFNN and RNN, proved to be very effective as they performed on par with or better than SOTA approaches.
In the future, we plan on investigating the sequence to sequence models such as RNN Seq2seq, Conv Seq2seq and Transformer. In another contribution of this work, we showed that diacritics can be integrated into other systems to attain enhanced versions in NLP tasks. We used MT as a case study and showed how our idea of ToD improved the results of the SOTA NMT system.

\section*{Acknowledgments}
We gratefully acknowledge the support of the Deanship of Research at the Jordan University of Science and Technology for supporting this work via Grant \#20180193 in addition to NVIDIA Corporation for the donation of the Titan Xp GPU that was used for this research.

\bibliography{emnlp-ijcnlp-2019}
\bibliographystyle{acl_natbib}

\clearpage

\appendix

\section{FFNN Models in Details}
\label{app:a}

This section discusses the details of the FFNN models.

\subsection{Basic Model}

\begin{figure}[!h]
    \centering
    \includegraphics[width=0.48\textwidth]{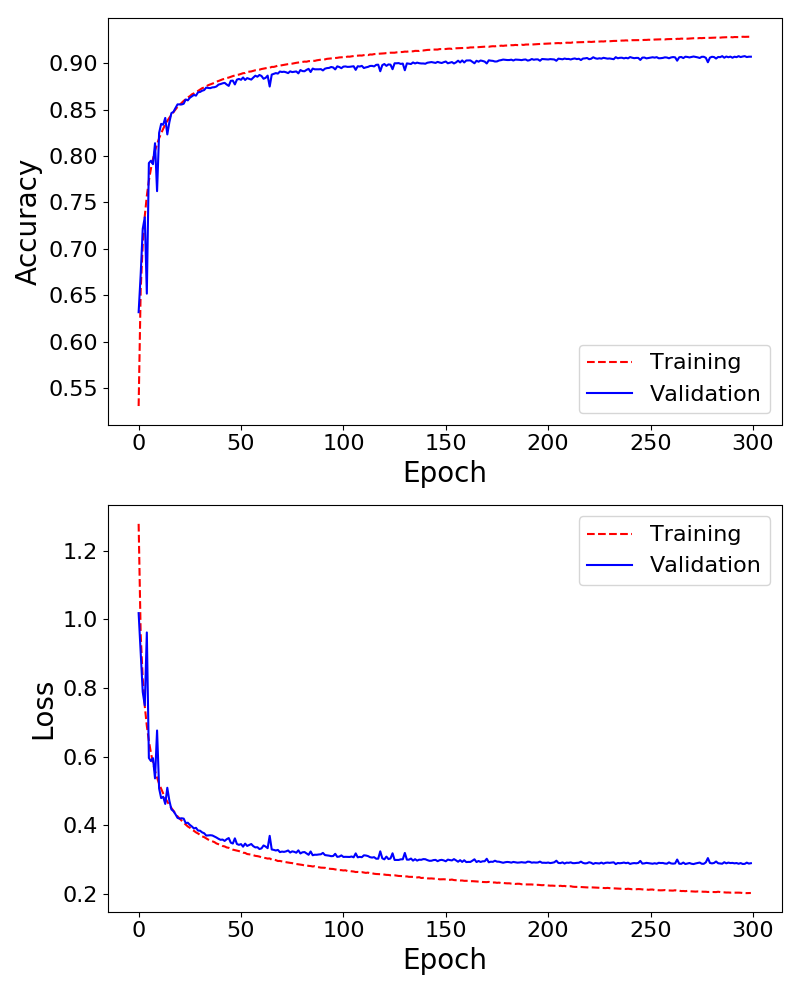}
    \caption{FFNN basic model training and validation accuracy and loss.}
    \label{fig:ffnn_basic_res}
\end{figure}

After a massive exploration for finding the best hyperparameters to structure the model (like the number of layers, number of neurons in each layer, and the activation function), the final structure is shown in Table~\ref{tab:ffnn_basic}. This model results in 92.87\%, 90.72\% and 90.67\% accuracies for training, validation, and testing datasets, respectively. Figure~\ref{fig:ffnn_basic_res} shows the loss and accuracy values on the training and validation datasets while training. The model is still able to slightly learn as well as generalize even after 300 epochs with no signs of overfitting.

\begin{table}[!h]
\centering
\caption{FFNN basic model structure}
\label{tab:ffnn_basic}
\begin{tabular}{|l|l|l|}
\hline
Layer Name & Neurons & Activation Func \\ \hline
Hidden 1 & 200 & ReLU \\ \hline
Hidden 2 & 500 & ReLU \\ \hline
Hidden 3 & 500 & ReLU \\ \hline
Hidden 4 & 450 & ReLU \\ \hline
Hidden 5 & 400 & ReLU \\ \hline
Hidden 6 & 400 & ReLU \\ \hline
Hidden 7 & 350 & ReLU \\ \hline
Hidden 8 & 300 & ReLU \\ \hline
Hidden 9 & 300 & ReLU \\ \hline
Hidden 10 & 250 & ReLU \\ \hline
Hidden 11 & 200 & ReLU \\ \hline
Hidden 12 & 200 & ReLU \\ \hline
Hidden 13 & 150 & ReLU \\ \hline
Hidden 14 & 100 & ReLU \\ \hline
Hidden 15 & 100 & ReLU \\ \hline
Hidden 16 & 50 & ReLU \\ \hline
Hidden 17 & 25 & ReLU \\ \hline
Output & 15 & Softmax \\ \hline
\multicolumn{3}{|l|}{Trainable Parameters: 1,501,115} \\ \hline
\end{tabular}
\end{table}

\subsection{100-Hot Model}

\begin{table}[h]
\centering
\caption{FFNN 100-Hot model structure}
\label{tab:ffnn_100}
\begin{tabular}{|l|l|l|}
\hline
Layer Name & Neurons & Activation Func \\ \hline
One Hot & N/A & N/A \\ \hline
Flatten & N/A & N/A \\ \hline
Dropout 1 (2.5\%) & N/A & N/A \\ \hline
Hidden 1 & 250 & ReLU \\ \hline
Dropout 2 (2.5\%) & N/A & N/A \\ \hline
Hidden 2 & 200 & ReLU \\ \hline
Dropout 3 (2.5\%) & N/A & N/A \\ \hline
Hidden 3 & 150 & ReLU \\ \hline
Dropout 4 (2.5\%) & N/A & N/A \\ \hline
Hidden 4 & 100 & ReLU \\ \hline
Dropout 5 (2.5\%) & N/A & N/A \\ \hline
Hidden 5 & 50 & ReLU \\ \hline
Dropout 6 (2.5\%) & N/A & N/A \\ \hline
Output & 15 & Softmax \\ \hline
\multicolumn{3}{|l|}{Trainable Parameters: 1,951,515} \\ \hline
\end{tabular}
\end{table}

\begin{figure}[h]
    \centering
    \includegraphics[width=0.48\textwidth]{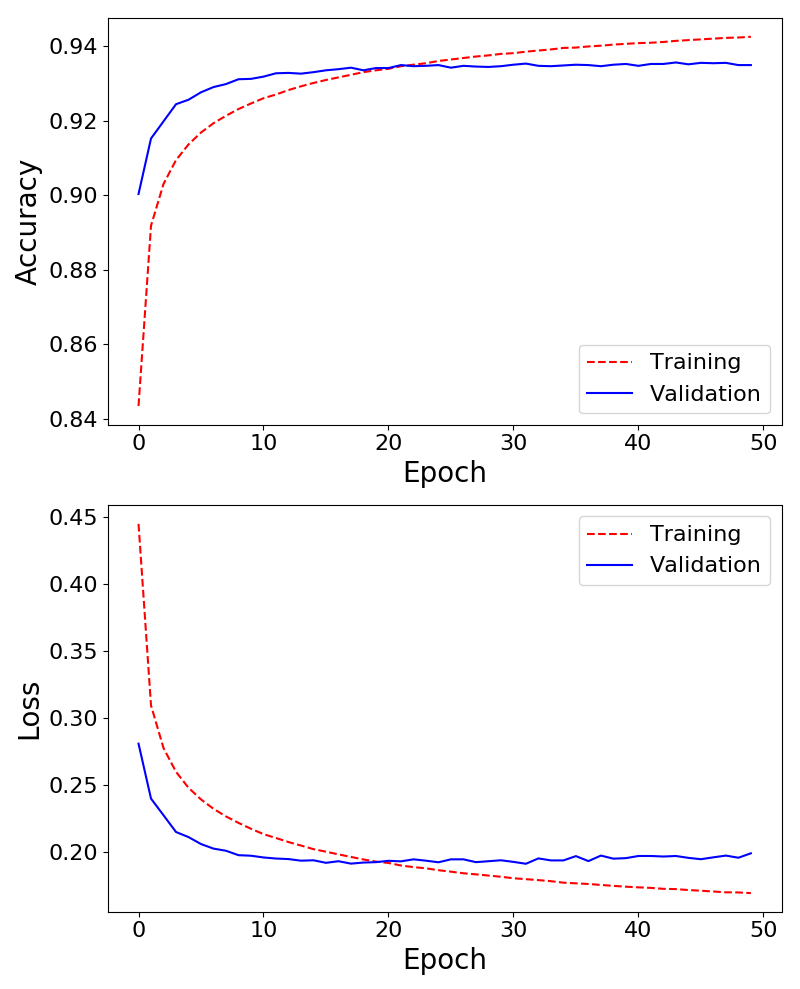}
    \caption{FFNN 100-Hot model training and validation accuracy and loss.}
    \label{fig:ffnn_100_res}
\end{figure}

Starting from the the basic model structure (Table~\ref{tab:ffnn_basic}), and by tuning the hyperparameters and using extra techniques, such as applying Dropout regularization, the model is able to learn better using the 100-hot representations. The model is structured as shown in Table~\ref{tab:ffnn_100}. This model results in 94.25\%, 93.49\% and 93.45\% accuracies for training, validation, and testing datasets, respectively. This is an improvement of 2.78\% on the test set accuracy compared to the basic model. Figure~\ref{fig:ffnn_100_res} shows the loss and accuracy values on the training and validation datasets while training.

\subsection{Embeddings Model}

\begin{table}[h]
\centering
\caption{FFNN Embeddings model structure}
\label{tab:ffnn_emb}
\begin{tabular}{|l|l|l|}
\hline
Layer Name & Neurons & Activation Func \\ \hline
Embedding (25) & N/A & N/A \\ \hline
Flatten & N/A & N/A \\ \hline
Dropout (10\%) & N/A & N/A \\ \hline
Hidden 1 & 250 & ReLU \\ \hline
Hidden 2 & 200 & ReLU \\ \hline
Hidden 3 & 150 & ReLU \\ \hline
Hidden 4 & 100 & ReLU \\ \hline
Hidden 5 & 50 & ReLU \\ \hline
Output & 15 & Softmax \\ \hline
\multicolumn{3}{|l|}{Trainable Parameters: 728,590} \\ \hline
\end{tabular}
\end{table}

\begin{figure}[h]
    \centering
    \includegraphics[width=0.48\textwidth]{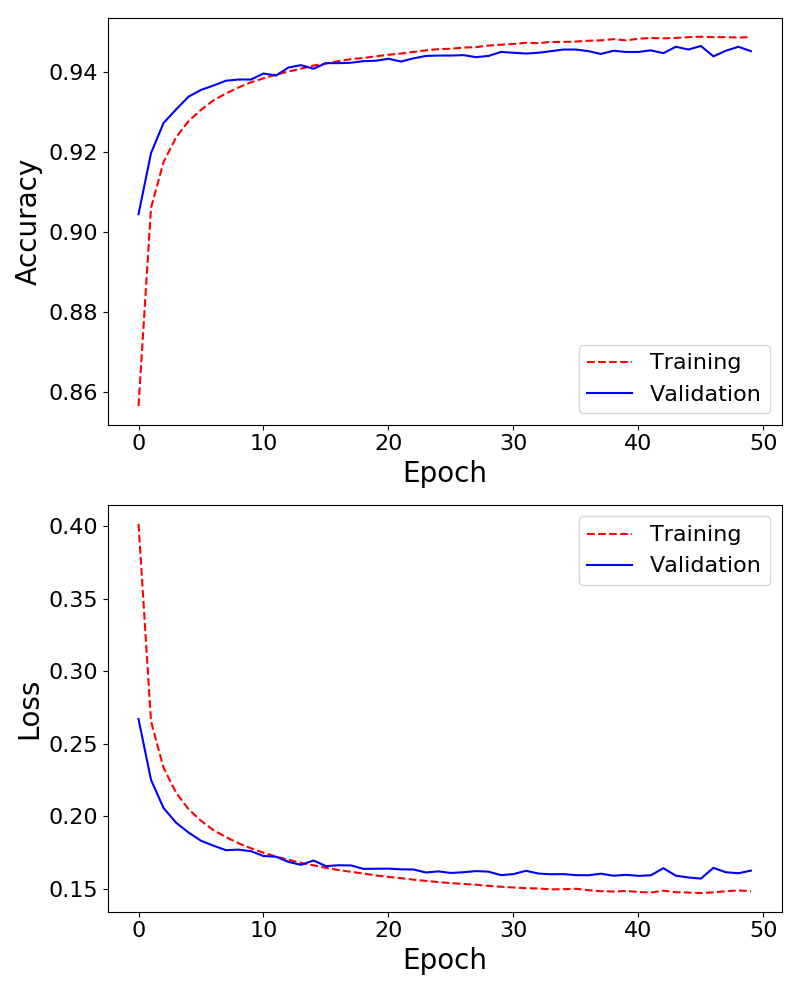}
    \caption{FFNN Embeddings model training and validation accuracy and loss.}
    \label{fig:ffnn_emb_res}
\end{figure}

Using a very similar structure as the 100-hot model structure (Table~\ref{tab:ffnn_100}), this model is structured as shown in Table~\ref{tab:ffnn_emb}. It achieves the best results compared to the basic and 100-hot models with 94.88\%, 94.53\% and 94.49\% accuracies for training, validation, and testing datasets, respectively. This model improves the accuracy by 1.04\% on the test set compared to the 100-hot model while reducing the number of trainable parameters (model size) by 51.46\% and 62.66\% compared to the basic and 100-hot models, respectively. Figure~\ref{fig:ffnn_emb_res} shows the loss and accuracy values on the training and validation datasets while training.

\begin{figure*}[h]
    \centering
    \includegraphics[width=0.79\textwidth]{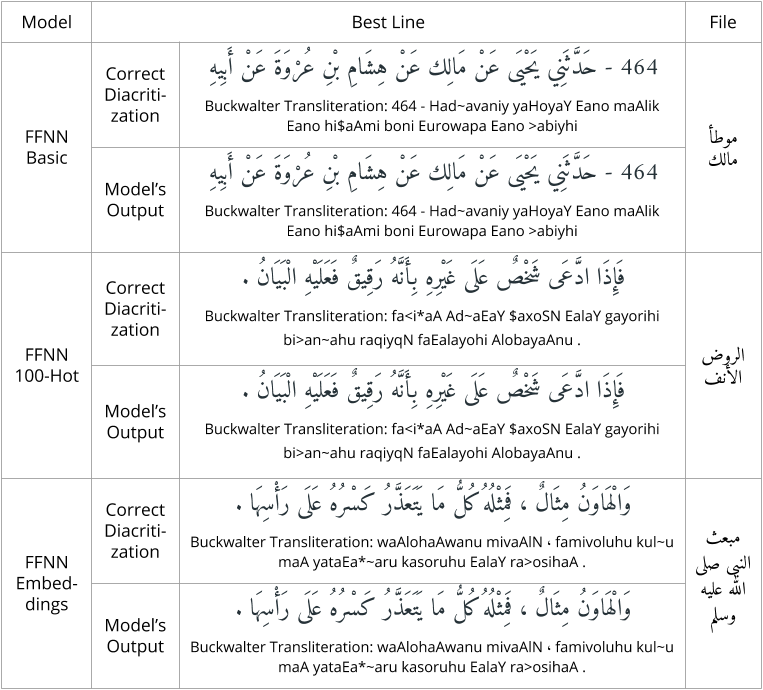}
    \caption{FFNN models good diacritization examples.}
    \label{fig:ffnn_good}
\end{figure*}

\begin{figure*}[h]
    \centering
    \includegraphics[width=0.78\textwidth]{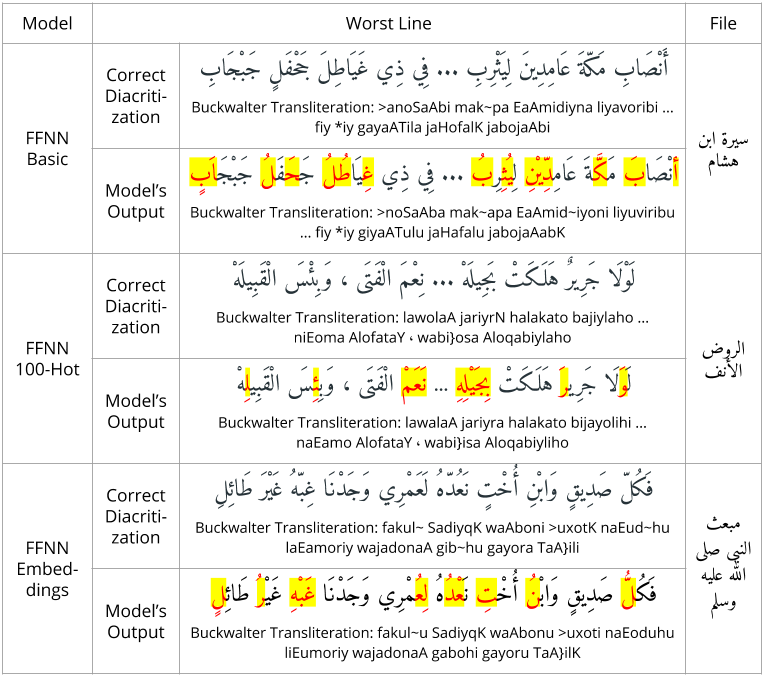}
    \caption{FFNN models bad diacritization examples.}
    \label{fig:ffnn_bad}
\end{figure*}

Figure~\ref{fig:ffnn_good} shows the best diacritization examples diacritized using each FFNN model, while Figure~\ref{fig:ffnn_bad} shows the worst diacritization examples. It is worth mentioning that the worst examples (listed in Figure~\ref{fig:ffnn_bad}) are from old Arabic poetry, which is very hard to diacritize flawlessly even for native speakers.

\section{RNN Models in Details}
\label{app:b}

\begin{figure*}[h]
    \centering
    \includegraphics[width=0.88\textwidth]{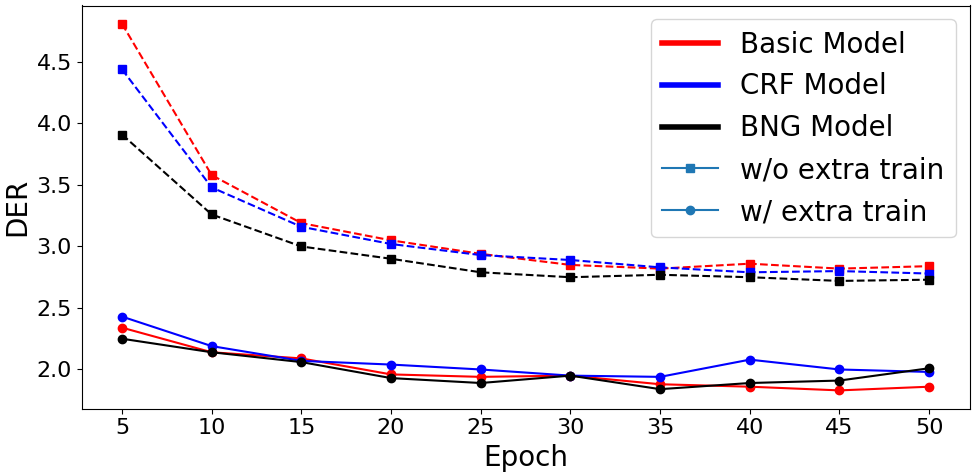}
    \caption{Recurrent models validation DER while training.}
    \label{fig:der_while_training}
\end{figure*}

This section provides details for the trained RNN models. First of all, Figure~\ref{fig:der_while_training} shows the validation DER of each model while training, reported every 5 epochs. This clarifies the importance of the dataset size, where any model significantly improves their DER when trained with the extra train dataset compared to any other model trained without it.

\begin{figure*}[h]
    \centering
    \includegraphics[width=0.88\textwidth]{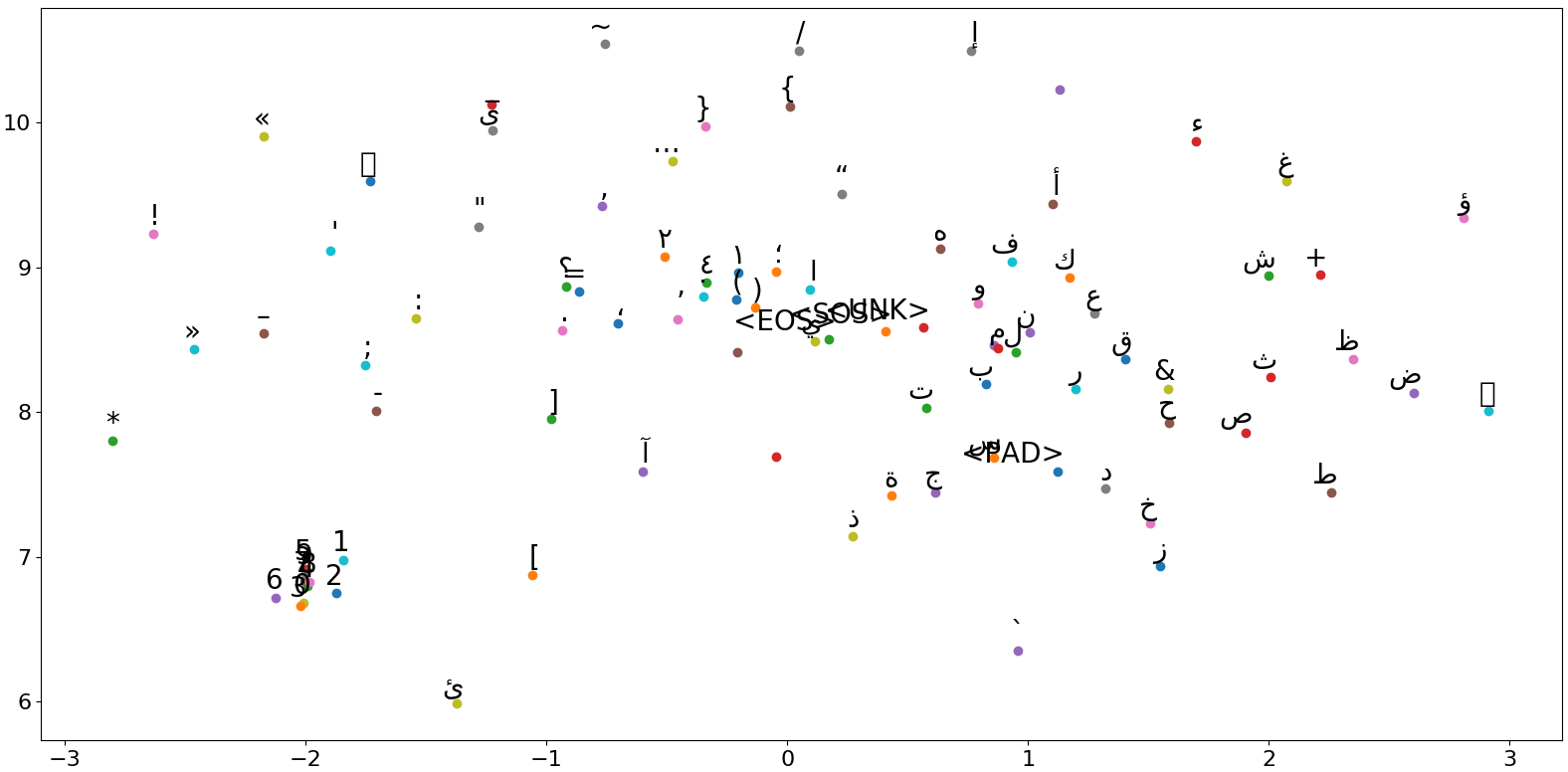}
    \caption{Embeddings plotted in 2D space.}
    \label{fig:vis}
\end{figure*}

Moreover, to explore the embeddings learnt by our best model, the weights vectors from the embeddings layer were extracted and reduced to 2 dimensions instead of 25 using t-SNE dimensionality reduction algorithm \cite{maaten2008visualizing}, then plotted in 2D space as shown in Figure~\ref{fig:vis}. The embeddings are able to capture meaningful information where digits appear together at the bottom-left, the majority of the punctuations appear at the middle and the top-left side, and finally, the Arabic letters appear at the right side.

\begin{figure*}[h]
    \centering
    \includegraphics[width=0.76\textwidth]{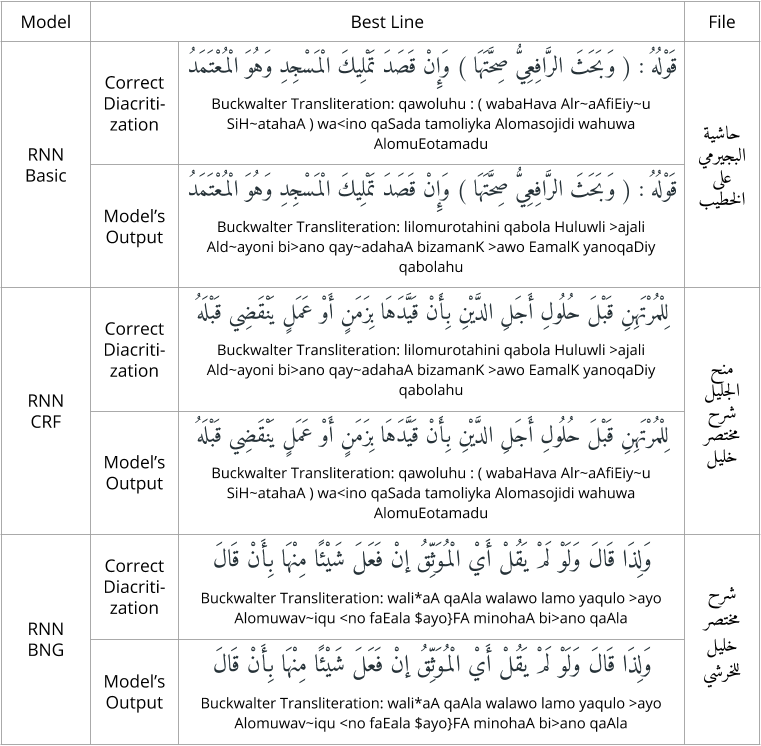}
    \caption{RNN models good diacritization examples.}
    \label{fig:rnn_good}
\end{figure*}

\begin{figure*}[h]
    \centering
    \includegraphics[width=0.76\textwidth]{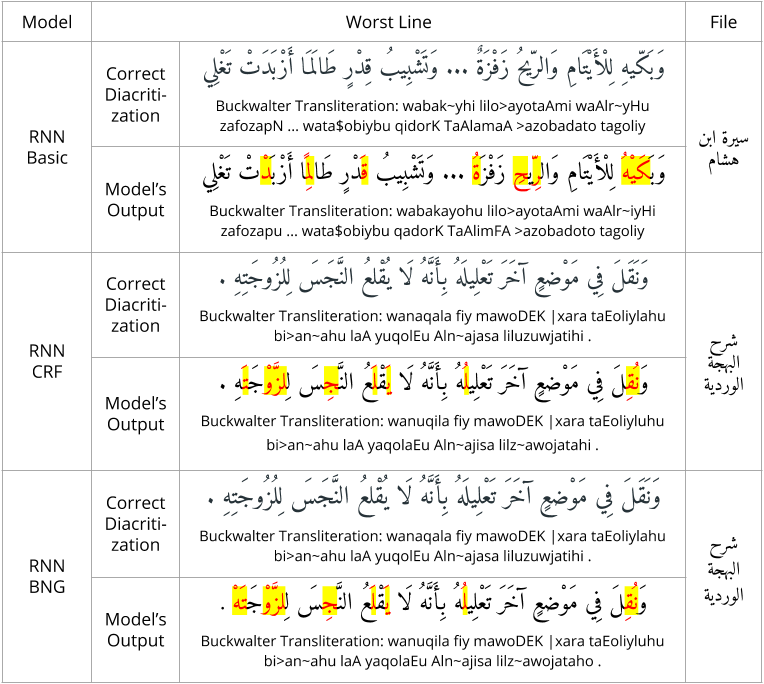}
    \caption{RNN models bad diacritization examples.}
    \label{fig:rnn_bad}
\end{figure*}

Figures \ref{fig:rnn_good} and \ref{fig:rnn_bad} show both best and worst examples from diacritizing using each RNN model. An important note is that the old Arabic poetry lines are no longer the majority in the worst examples, in contrast to the FFNN models.

\begin{figure*}[h]
    \centering
    \includegraphics[width=0.85\textwidth]{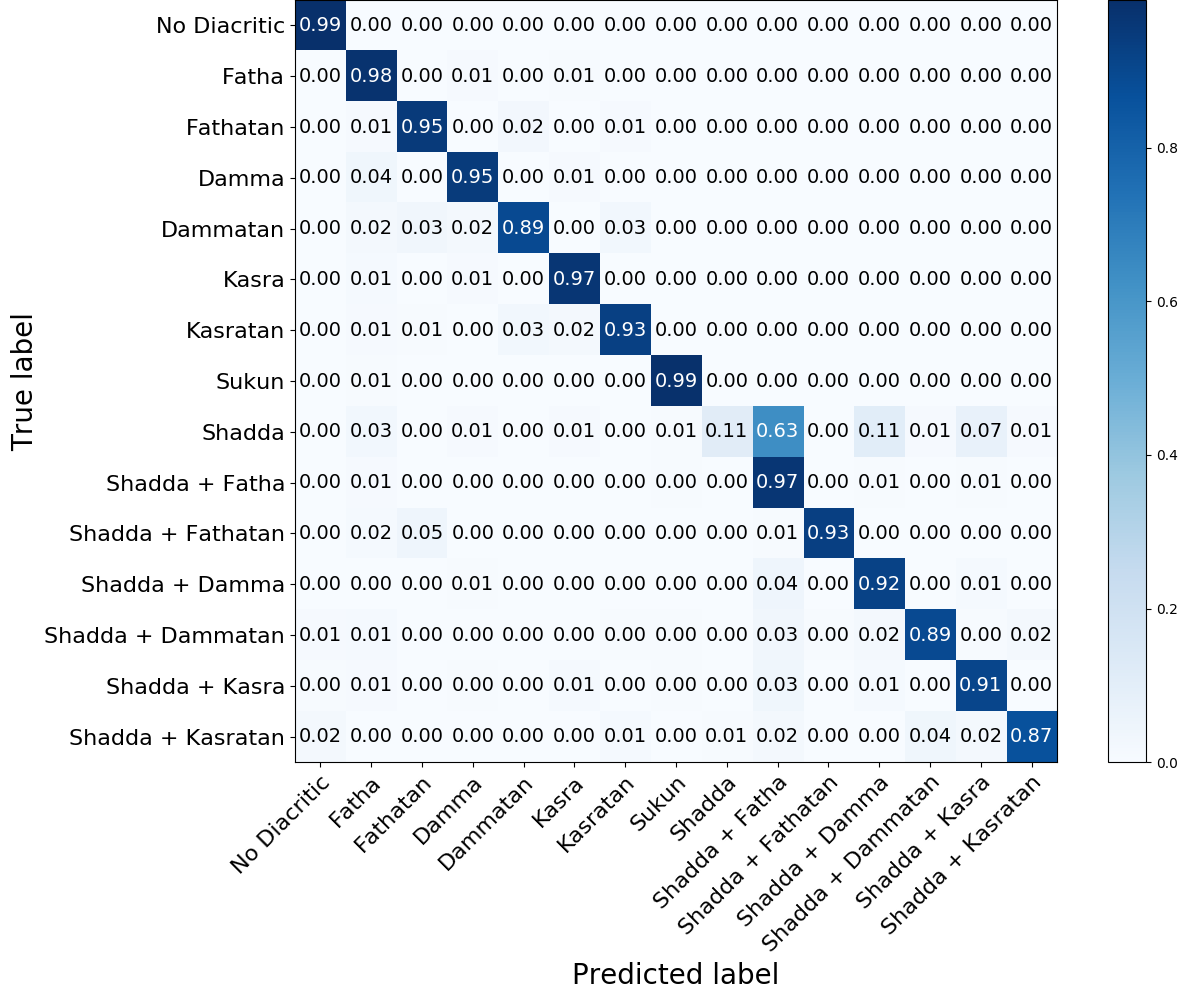}
    \caption{Without extra train confusion matrix for the best BNG model.}
    \label{fig:without_extra_conf}
\end{figure*}

\begin{figure*}[h]
    \centering
    \includegraphics[width=0.85\textwidth]{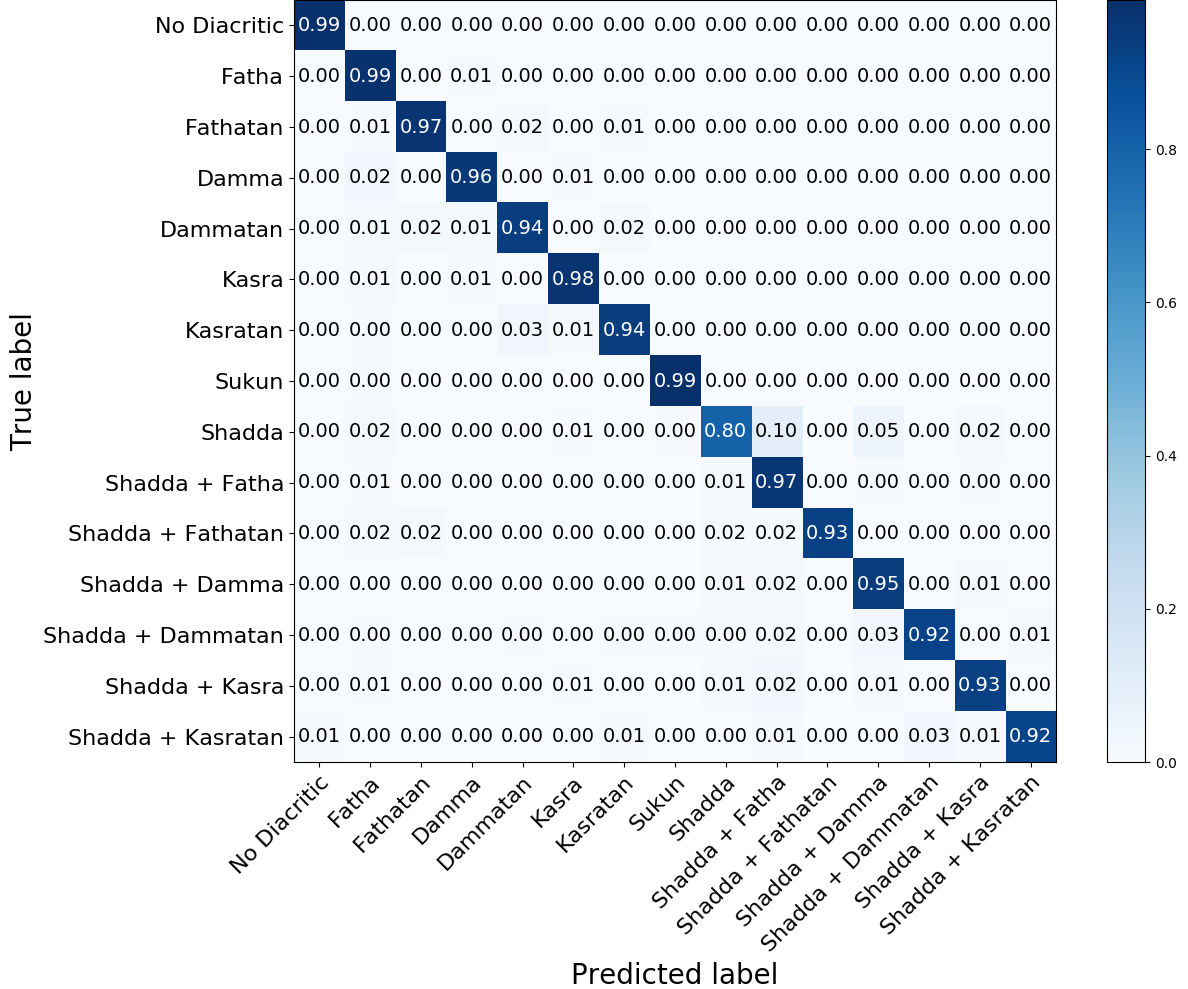}
    \caption{With extra train confusion matrix for the best BNG model.}
    \label{fig:with_extra_conf}
\end{figure*}

Finally, Figures \ref{fig:without_extra_conf} and \ref{fig:with_extra_conf} shows the confusion matrices related to our best model when trained without and with the extra train dataset, respectively. By comparing them, it is easy to see that the Shadda class is the worst one in both cases. However, the case with the extra train dataset shows dramatic improvement in this class, as well as other classes like Shadda + another diacritic and the Dammatan. A justification for this improvement is that there is a larger number of examples in the extra train dataset related to these classes as shown in Table~\ref{tab:num_examples}. Another insight can be concluded from the confusion matrices is that the model usually misclassifies the Shadda class as Shadda + another diacritic class due to different diacritization conventions, which in many cases would be a grammatically correct guess.

%\clearpage

\begin{table*}
\centering
\caption{Number of examples for each class}
\label{tab:num_examples}
\begin{tabular}{c|c|c|c|c|c|c|}
\cline{2-7}
                                         & Train  & Valid & Test & Extra Train & Total   & \%    \\ \hline
\multicolumn{1}{|c|}{No Diacritic}       & 4,366K & 213K  & 222K & 46,647K     & 51,449K & 38.87 \\ \hline
\multicolumn{1}{|c|}{Fatha}              & 2,932K & 144K  & 150K & 31,287K     & 34,514K & 26.07 \\ \hline
\multicolumn{1}{|c|}{Fathatah}  & 58K & 3K & 3K & 626K & 691K & 00.52 \\ \hline
\multicolumn{1}{|c|}{Damma}              & 812K   & 39K   & 41K  & 8,648K      & 9,539K  & 07.20 \\ \hline
\multicolumn{1}{|c|}{Dammatan} & 58K & 3K & 3K & 622K & 686K & 00.51 \\ \hline
\multicolumn{1}{|c|}{Kasra}              & 1,265K & 62K   & 64K  & 13,533K     & 14,924K & 11.27 \\ \hline
\multicolumn{1}{|c|}{Kasratan} & 88K & 4K & 4K & 941K & 1,037K & 00.78 \\ \hline
\multicolumn{1}{|c|}{Sukun}              & 1,230K & 60K   & 63K  & 13,135K     & 14,487K & 10.94 \\ \hline
\multicolumn{1}{|c|}{Shaddah} & 6K & 254 & 471 & 66K & 73K & 00.05 \\ \hline
\multicolumn{1}{|c|}{Shaddah + Fatha}    & 300K   & 15K   & 15K  & 3,202K      & 3,532K  & 02.66 \\ \hline
\multicolumn{1}{|c|}{Shaddah + Fathatah} & 3K & 189 & 132 & 36K & 40K & 00.03 \\ \hline
\multicolumn{1}{|c|}{Shaddah + Damma} & 43K & 2K & 2K & 463K & 511K & 00.38 \\ \hline
\multicolumn{1}{|c|}{Shaddah + Dammatan} & 5K & 238 & 222 & 51K & 56K & 00.04 \\ \hline
\multicolumn{1}{|c|}{Shaddah + Kasra} & 64K & 3K & 3K & 679K & 749K & 00.56 \\ \hline
\multicolumn{1}{|c|}{Shaddah + Kasratan} & 6K & 298 & 273 & 63K & 69K & 00.05 \\ \hline
\end{tabular}
\end{table*}

% \begin{table}[]
% \begin{tabular}{llllll}
%                   & Train   & Valid  & Test   & Extra Train & Total    \\
% No Diacritic       & 4366391 & 213095 & 222172 & 46647263    & 51448921 \\
% Fatha              & 2932361 & 143655 & 150213 & 31287371    & 34513600 \\
% Fathatah           & 58104   & 2981   & 3200   & 626375      & 690660   \\
% Damma              & 811590  & 38771  & 41070  & 8647647     & 9539078  \\
% Dammatan           & 58115   & 2850   & 2927   & 621967      & 685859   \\
% Kasra              & 1264779 & 61644  & 64448  & 13533150    & 14924021 \\
% Kasratan           & 87730   & 4352   & 4349   & 940701      & 1037132  \\
% Sukun              & 1229871 & 60007  & 62656  & 13134500    & 14487034 \\
% Shaddah            & 6240    & 254    & 471    & 66118       & 73083    \\
% Shaddah + Fatha    & 299828  & 14540  & 15274  & 3202424     & 3532066  \\
% Shaddah + Fathatah & 3492    & 189    & 132    & 36469       & 40282    \\
% Shaddah + Damma    & 43329   & 2169   & 2176   & 463292      & 510966   \\
% Shaddah + Dammatan & 4757    & 238    & 222    & 51039       & 56256    \\
% Shaddah + Kasra    & 63514   & 3142   & 3119   & 679382      & 749157   \\
% Shaddah + Kasratan & 5841    & 298    & 273    & 62737       & 69149
% \end{tabular}
% \end{table}

\end{document}